\documentclass{ifacconf}
\usepackage{amsmath,amssymb,amsfonts}
\usepackage{graphicx}      
\usepackage{natbib}        
\usepackage{xcolor}
\usepackage{tabu}
\usepackage{booktabs}
\usepackage{tikz}
\usepackage{pgfplots}
\usepackage{xurl}

\usepackage{fancyhdr}
\fancypagestyle{firstpage}{%
  \lhead{Final version, Presented at IFAC World Congress 2023, Yokohama, Japan}
  \rhead{}
}
\definecolor{color0}{rgb}{0.8235,0,0} 
\definecolor{color1}{rgb}{0.07843,0.549,0.07843} 
\definecolor{color2}{rgb}{0,0,1} 
\definecolor{color3}{rgb}{1,0.5137,0.4824} 
\definecolor{color4}{rgb}{0.5098,0.8588,0.1961} 
\definecolor{color5}{rgb}{0.4314,0.7451,0.9804} 
\definecolor{color6}{rgb}{0.3451,0.3451,0.3451} 
\definecolor{color7}{rgb}{0.6863,0.6863,0.6863} 
\definecolor{color8}{rgb}{0,0,0} 
\definecolor{color9}{rgb}{0.75,0.25,0} 
\definecolor{color10}{rgb}{0,0.8,0} 
\definecolor{color11}{rgb}{0.44, 0, 0.8 } 

\newcommand{\plotLine}{1.25pt}
\newcommand{\plotDotted}{1.5pt}

\newcommand{\tr}{\text{tr}}
\newcommand{\so}{\mathfrak{so}(3)}
\newcommand{\se}{\mathfrak{se}(3)}
\newcommand{\SO}{SO(3)}
\newcommand{\SE}{SE(3)}

\newcommand{\hmap}{\textit{hat-map}}
\newcommand{\vmap}{\textit{vee-map}}
	
\newcommand{\eg}{e_{\g}}

\newcommand{\pvec}{p}
\newcommand{\pvecd}{p_d}

\newcommand{\ev}{e_V}
\newcommand{\evn}{\|e_V\|_2}

\newcommand{\g}{g}

\newcommand{\gd}{g_d}

\begin{document}
\begin{frontmatter}

\title{\vspace{-10pt}Geometric Impedance Control on $\SE$ for Robotic Manipulators} 



\author[First]{Joohwan Seo} 
\author[First]{Nikhil Potu Surya Prakash} 
\author[Second]{Alexander Rose}
\author[third]{Jongeun Choi}
\author[First]{Roberto Horowitz}

\address[First]{University of California, Berkeley, CA, 94720 USA}
\address[Second]{Leibniz University Hannover, Germany}
\address[third]{Yonsei University, Seoul, Republic of Korea
\\ e-mail: \{joohwan\_seo, nikhilps, rose1993, horowitz\}@berkeley.edu, 
jongeunchoi@yonsei.ac.kr
}

\begin{abstract} 
After its introduction, 
impedance control has been utilized as a primary control scheme for robotic manipulation tasks that involve interaction with unknown environments.
While impedance control has been extensively studied, the geometric structure of  $\SE$ for the robotic manipulator itself and its use in formulating a robotic task has not been adequately addressed.
%
In this paper, we propose a differential geometric approach to impedance control.
%
%
Given a left-invariant error metric in $\SE$, the corresponding error vectors in position and velocity are first derived.
We then propose the impedance control schemes that adequately account for the geometric structure of the manipulator in $\SE$ based on a left-invariant potential function.
%
%
The closed-loop stabilities for the proposed control schemes are verified using Lyapunov function-based analysis.
%
The proposed control design clearly outperformed a conventional impedance control approach when tracking challenging trajectory profiles.
\end{abstract}


\end{frontmatter}
\thispagestyle{firstpage}
\section{Introduction}\label{Sec:1}
Impedance control was first proposed to solve the manipulator's end-effector positioning problem \cite{hogan1985impedance}.
Since then, often combined with an operational space frame formulation \cite{khatib1987unified}, it has been the main framework for controlling the end-effector position in the Cartesian space, particularly on tasks that involve interaction with the environment, and is thereby widely utilized in robotic manipulation \cite{abu2020variable}. 

The end-effector impedance can be configured to ensure safety in interaction-rich tasks. Particularly, impedance control enables an engineering trade-off between positioning accuracy requirements and robustness against unmodeled interactions
\cite{chen2013robotic, khan2010safe, ott2008passivity}. To better deal with the unknown environment, impedance control is often combined with learning algorithms; adaptive control approaches such as \cite{li2010learning, duan2018adaptive},
behavior cloning approaches including \cite{rozo2016learning, abu2018force}
and reinforcement learning approaches \cite{zhang2021learning, martin2019variable, luo2019reinforcement}.

Tracking errors in the Cartesian space impedance control are often calculated separately as translation and orientation errors in the spatial frame. 
In \cite{caccavale1999six}, representations for the orientation error utilizing rotation matrices are proposed.
The concept of using the geometry of $\SO$ to define orientation errors is widely employed, e.g., \cite{robosuite2020, ochoa2021impedance}. However, to date, most schemes utilized for separating translation and orientation errors do not adequately consider the geometry of $\SE$, even though the end-effector lies in $\SE$.

In impedance control, three major components address the dynamic behavior between the manipulator and the environment: mass, spring, and damper.
Among these components, a spatial spring defined in $\SE$ was utilized in robotic applications by \cite{stramigioli2001variable} and also in the control of Unmanned Aerial Vehicles (UAV) by \cite{rashad2019energy}. The wrenches from the spatial springs and linear dampers are utilized as the control inputs. Based on a defined spatial spring, the impedance control was derived in \cite{fasse1997spatial} in the spatial frame. Although the manifold structure of $\SE$ is considered in these works,  results involving time-varying trajectories and a  stability analysis were not provided. 

Thus, to the best of our knowledge, the intrinsic manifold and geometrical structure of impedance control from the control perspective has not yet been rigorously examined. 
For example, in general, desired and current Cartesian velocities are directly compared even though they are not in the same tangent spaces of the $\SE$ manifold \cite{li2010learning,khan2010safe}.
Additionally, since there is no bi-invariant error metric in $\SE$ \cite{park1995distance}, \cite{bullo1995proportional}, the position error vectors need to be defined so that they are consistent with the error metric that has been selected. 
For example, utilizing spatial frame-based error vectors for left-invariant metrics is inconsistent with its geometry, and the performance of the control system can be increased when  error vectors are properly defined.

In this paper, we revisit  impedance control for robotic manipulators in the context of differential geometry. We first define position and velocity errors in the end-effector body frame lying on its tangent space, followed by two versions of a  geometric impedance control law, built upon  results from \cite{bullo1999tracking}. It is shown with a numerical simulation example that our proposed control scheme outperforms conventional non-geometric impedance control approaches in trajectory tracking. 

This paper can be considered as an extension of \cite{lee2010geometric} from $\SO$ to $\SE$ and from UAVs to robotic manipulators, and is closely related to the geometric control proposed by \cite{bullo1999tracking} in the context of Cartesian operation space impedance control.
Compared to \cite{lee2010geometric}, which utilizes a scalar gain, we define the rotational potential function utilizing a matrix impedance gain, which is important in the framework of variable impedance control. We provide a justification for formalizing a left-invariant potential function, which is one of the several  such functions suggested by \cite{bullo1999tracking}. In addition, we present complete stability analyses for both controllers, which have been previously omitted for $\SE$, while minimizing the use of advanced differential geometric mathematical constructs and notations.
We emphasize that using correct error vectors consistent with the task defined in $\SE$ will be critical to learning the required variable impedance control gains in many learning assembly operations since this will minimize the detrimental effects during learning caused by erroneous manipulator behavior due to inconsistent error tracking. Furthermore, the left-invariant error vector can be useful in encoding manipulation tasks lying in $\SE$.

The paper is organized as follows:
Section.~\ref{Sec:2}, presents a brief review of Lie groups in $\SO$ and $\SE$ and their respective $\so$ and $\se$ algebras, as well a brief review of its use in forward kinematics and  manipulator dynamics.
In Section~\ref{Sec:3}, the error function in $\SE$ proposed in this paper and its corresponding error vectors are derived, as well as  geometrically consistent impedance control laws. Stability analyses of the corresponding  closed-loop systems are provided as well. 
Section~\ref{Sec:4} shows the simulation results of the proposed approach. Benchmark results are also presented in comparison to the conventional impedance controller.
Finally,  concluding remarks and future work are discussed in Section~\ref{Sec:5}.
\vspace{-5pt}

\section{Manipulator Dynamic Model} \label{Sec:2}
\subsection{Lie group and Lie algebra of $\SO$ and $\SE$} \vspace{-5pt}
In this subsection, we briefly review the contents of group structure and its geometry.
For the detailed description regarding Lie group \& Lie algebra and their applications to robotics, the reader is referred to \cite{murray1994mathematical} and \cite{lynch2017modern}.
The configuration of a robotic manipulator's end-effector is defined by its position and orientation; the configuration manifold lies in the special Euclidean group $\SE$. 
The homogeneous representation of the $\SE$ configuration manifold is given by
\begin{equation}\begin{aligned}
	\label{eq:homo}
    g = \begin{bmatrix}
    R & \pvec \\ 0 & 1
    \end{bmatrix}\!\in\!\SE.
\end{aligned}\end{equation}
where $R\!\in\!\SO$, $\pvec\!\in\!\mathbb{R}^3$, and $\SO\!=\!\{R\!\in\!\mathbb{R}^{3\times3}| R^TR = R R^T = I, \text{det}(R)\!=\!1\}$.

The Lie algebra of $\SO$, denoted by $\so$, defines  skew-symmetric matrices. 
The hat map for $\so$ is defined by $\widehat{(\cdot)}:\mathbb{R}^3\!\to\!\so$, which maps an angular velocity vector $\omega \in \mathbb{R}^3$ to its skew-symmetric cross product matrix i.e., $\hat{\omega}\!=\!-\hat{\omega}^T$ for $\omega\!\in\!\mathbb{R}^3$. Thus, for any vector $v \in \mathbb{R}^3$, $\hat{\omega} v = \omega \times v$.
Whereby an element in $\so$ can be mapped form a vector in $\mathbb{R}^3$ using the $\hmap\; \widehat{(\cdot)}$: $\so\!\to\!\mathbb{R}^3$.
The  inverse of the hat map, $\widehat{(\cdot)}$, is the $\vmap\; (\cdot)^\vee$: $\so\!\to\!\mathbb{R}^3$.
Similarly to $\SO$, the Lie algebra of $\SE$ is denoted by $\se$.
The $\hmap$ for $\se$ is defined as $\widehat{(\cdot)}$: $\mathbb{R}^6\!\to\!\se$.
The inverse map of hat map is defined by $\vmap\; (\cdot)^\vee$: $\se\!\to\!\mathbb{R}^6$. Thus,  
\begin{equation}\begin{aligned}
    \hat{\xi} \!=\! \begin{bmatrix}
    \hat{\omega} & v \\ 0 & 0
    \end{bmatrix}\!\in\!\se, \; \forall \xi \!=\! \begin{bmatrix} v\\ \omega \end{bmatrix}\!\in\!\mathbb{R}^6, \; v,\omega\!\in\!\mathbb{R}^3, \;\hat{\omega} \!\in\! \so.
\end{aligned}\end{equation}
%


\subsection{Forward Kinematics} \vspace{-5pt}
We begin this subsection by noting that our interest is in robotic manipulators with revolute joints.
%
We also use the matrix exponential given as
\begin{equation}\begin{aligned} \label{eq:matrix_exp}
   e^{A}=   \sum_{k=0}^{\infty} \tfrac{1}{k!}A^k, \quad A\!\in\!\mathbb{R}^{n\times n}
\end{aligned}\end{equation}
to describe the kinematics of the manipulator as in \cite{murray1994mathematical}:
\begin{equation}\begin{aligned}
    g(q) = e^{\hat{\xi}_1 q_1} e^{\hat{\xi}_2 q_2} \cdots e^{\hat{\xi}_n q_n}g(0) = \begin{bmatrix}
    R(q) & \pvec(q) \\ 0 & 1
    \end{bmatrix}\!\in\!\SE
\end{aligned}\end{equation}
where $g(q)$ is the homogeneous representation~\eqref{eq:homo} of the end-effector in the spatial frame, $\xi_i\!\in\!\mathbb{R}^6$ and $q_i\!\in\!\mathbb{R}$ are a twist represented in the spatial frame and a joint angle of the $i\textsuperscript{th}$ joint, respectively, for $i\!=\!1,2,\dots, n$, and $n$ is the number of joints.
In addition, $g(0)$ is an initial configuration.
Since we are dealing with revolute joints, the joint generalized coordinate vector is $q\!=\![q_1,\cdots,q_n]^T\!\in\!\mathcal{S}$ with $\mathcal{S} \triangleq \mathbb{S}^1\times \cdots \times \mathbb{S}^1$ (repeated by $n$ times).

The velocity of the end-effector in the body frame,
$\left.V^b \in \mathbb{R}^{6}\right.$, can be calculated using following formula:
\begin{equation}\begin{aligned}
    \hat{V}^b = \widehat{\begin{bmatrix}
    v^b \\ \omega^b
    \end{bmatrix}} = 
    g^{-1} \dot{g}
\end{aligned}\end{equation}
i.e., $\dot{g} = g \hat{V}^b$.
The velocity $V^b$ can also be computed using the body Jacobian matrix $J_b(q)$ as follows:
\begin{equation}\begin{aligned}\label{eq:velo_jacobian}
    V^b = J_b(q) \dot{q}.
\end{aligned}\end{equation}
For the details about $J_b(q)$, we refer to Chap 5.1 of \cite{lynch2017modern}.

\subsection{Manipulator Dynamics} \vspace{-5pt}
We consider the well-known manipulator dynamics equation of motion 
\begin{equation}\begin{aligned} \label{eq:robot_dynamics}
    M(q)\ddot{q} + C(q,\dot{q})\dot{q} + G(q) = T + T_e,
\end{aligned}\end{equation}
where $M(q)\!\in\!\mathbb{R}^{n\times n}$ is the symmetric positive definite inertia matrix, $C(q,\dot{q})\!\in\!\mathbb{R}^{n \times n}$ is a Coriolis matrix, \hfill \break
$G(q)\!\in\!\mathbb{R}^n$ is a moment term due to gravity, $T\!\in\!\mathbb{R}^n$ is a control input given in torque, and $T_e\!\in\!\mathbb{R}^n$ is an external disturbance in torque. 
The $(r,s)$ element of the matrix $C(q,\dot{q})$ is 
\begin{equation}\begin{aligned}
    C_{rs}(q,\dot{q}) = \dfrac{1}{2}\sum_{t=1}^n\left[
    \dfrac{\partial M_{rs}}{\partial q_t}\dot{q}_t +
    \dfrac{\partial M_{tr}}{\partial q_s}\dot{q}_t -
    \dfrac{\partial M_{ts}}{\partial q_r}\dot{q}_t
    \right]\!,
\end{aligned}\end{equation}
see \cite{li1999passive}.
The Coriolis matrix calculated in this way satisfies the property that $\dot{M}\!-\!2C$ is skew-symmetric.

In the field of impedance control combined with operational space formulation, it is well known from \cite{khatib1987unified} that the robot dynamics \eqref{eq:robot_dynamics} can be rewritten as 
\begin{align} \label{eq:robot_dynamics_eef}
    \hspace{-5pt}\tilde{M}(q)\dot{V}^b &+ \tilde{C}(q,\dot{q})V^b + \tilde{G}(q) = \tilde{T} + \tilde{T}_e, \text{ where}\\
    \tilde{M}(q)&= J_b(q)^{-T} M(q) J_b(q)^{-1}, \nonumber\\
    \tilde{C}(q,\dot{q})&=J_b(q)^{-T}(C(q,\dot{q}) \!-\! M(q) J_b(q)^{-1}\dot{J})J_b(q)^{-1}, \nonumber\\
    \tilde{G}(q)&= J_b(q)^{-T} G(q),\; \tilde{T}= J_b(q)^{-T} T,\; \tilde{T}_e= J_b(q)^{-T} T_e.\nonumber
\end{align}
For compactness we use notation $\left(A^{-1}\right)^{T}\!=\!A^{-T}$.
Furthermore, we denote $\tilde{M}(q)$ as $\tilde{M}$, $\tilde{C}(q,\dot{q})$ as $\tilde{C}$ and $\tilde{G}(q)$ as $\tilde{G}$.\
We also note that the manipulator dynamics of the workspace is based on the following assumption.
\begin{assum}\label{assum:1}
    The end-effector lies in a region $\left.D \subset \SE\right.$ such that $J_b$ is full-rank, i.e., non-singular.
\end{assum}

\section{Geometric Impedance Control on $\SE$} \label{Sec:3}
\subsection{Error Vector Derivation} \vspace{-5pt}
In \cite{lee2010geometric}, the authors suggest that an error function in $\SO$ can be chosen as 
\begin{equation}\begin{aligned} \label{eq:metric_original}
    \psi_R(R,R_d) = \tr(I - R_d^T R),
\end{aligned}\end{equation}
where $R\!\in\!\SO$ and $R_d\!\in\!\SO$ respectively represent the actual and desired rotation matrices.
%
In $\SO$, the distance metric in \eqref{eq:metric_original} can also be defined using the Frobenius norm $\|\cdot\|_F$, as proposed in \cite{huynh2009metrics},
\begin{equation}\begin{aligned}\label{eq:metric_2}
    \psi_R(R,R_d)&= \tfrac{1}{2}\|I - R_d^T R\|^2_F\\
    &= \tfrac{1}{2}\tr\left((I - R_d^TR)^T(I - R_d^TR)\right)\!.
\end{aligned}\end{equation}
%
The formulation in \eqref{eq:metric_2} can be extended to general matrix groups.
Therefore, we utilize the Frobenius norm to define an error function in $\SE$ as:
\begin{equation}\begin{aligned} \label{eq:err_fun}
    \Psi(\g,\gd) = \tfrac{1}{2}\|I - \gd^{-1}\g\|^2_F,
\end{aligned}\end{equation}
where $g \in \SE$ and $g_d \in \SE$ respectively represent
the actual and desired configurations.
Utilizing the homogeneous representation \eqref{eq:homo}, it follows that
\begin{equation}\begin{aligned}
    \g = \begin{bmatrix}
    R & \pvec \\
    0 & 1
    \end{bmatrix}\!, \quad \gd = \begin{bmatrix}
    R_d & \pvecd \\
    0 & 1
    \end{bmatrix}\!, \quad \g^{-1} = \begin{bmatrix}
    R^T & - R^T \pvec\\
    0 & 1
    \end{bmatrix}\!.
\end{aligned}\end{equation}
This leads to the equivalent formulation for the error function $\Psi(\g,\gd)$ in  \eqref{eq:err_fun}:
\begin{equation}\begin{aligned} \label{eq:err_fun_SE3}
    \Psi(\g,\gd) = \tr(I - R_d^TR) + \tfrac{1}{2}(\pvec - \pvecd)^T(\pvec - \pvecd).
\end{aligned}\end{equation}
This error function is related to the squared of the distance metric in $\SE$ proposed in \cite{park1995distance}, in the sense that the error function in \cite{park1995distance} is also defined as a weighted summation of metrics in $\SO$ and the Euclidean space. However, in  \cite{park1995distance} the bi-invariant metric $\| \log(R^TR_d)\|$ is utilized for $\SO$.
In addition, the error function derived herein is the non-weighted version of one of the error functions proposed by \cite{bullo1999tracking}. We also emphasize that, by using the Frobenius norm-induced error function, we could justify the selection of the weighted error function in \cite{bullo1999tracking}.
It is important to note that the $\SE$ error function \eqref{eq:err_fun_SE3} is left-invariant i.e.,
\begin{equation}\label{eq:left_invariant}
    \begin{aligned}
        \Psi(g_l\g,g_l\gd) = \tfrac{1}{2}\| I - \gd^{-1}g_l^{-1}g_l\g\|^2_F = \Psi(g,g_d),
    \end{aligned}
\end{equation}
where $g_l\!\in\!\SE$ is an arbitrary left translation.
On the other hand, the right translation of the error function does not satisfy equation \eqref{eq:err_fun_SE3}.
Applying the right translations reads
\begin{equation}\label{eq:right_invariant}
    \begin{aligned}
        \Psi(\g g_r,\gd g_r) = \tfrac{1}{2}\| I - g_r^{-1}\gd^{-1}\g g_r\|^2_F \neq \Psi(g,g_d),
    \end{aligned}
\end{equation}
where $g_r\!\in\!\SE$ is an arbitrary right translation,
which confirms that the error function is left-invariant. As suggested by \cite{park1995distance}, it is natural to use body-frame coordinates for  left-invariant error metrics in $\SE$.

To calculate the error vector in $\SE$, we  perturb a configuration $\g \in \SE$ with a right translation, as was done as in \cite{lee2010geometric} for $\SO$.
\begin{equation}\begin{aligned}
    \g_{\hat{\eta}}= g e^{\hat{\eta}\epsilon}\!, \quad \eta=\begin{bmatrix}\eta_1\\ \eta_2\end{bmatrix}\!\in\!\mathbb{R}^6 \implies \hat{\eta}\!\in\!\se, \;\epsilon\!\in\!\mathbb{R},
\end{aligned}\end{equation}
where $\eta_1, \eta_2\!\in\!\mathbb{R}^3$ refer to translational and rotational components, respectively.
%
Therefore, the configuration variation  can be represented by the homogeneous transformation matrix
\begin{equation}\begin{aligned}
    \delta g_{\hat{\eta}} = g \hat{\eta} \implies \begin{bmatrix}
    \delta R & \delta \pvec \\
    0 & 0 
    \end{bmatrix}  = \begin{bmatrix}
    R\hat{\eta}_2 & R \eta_1 \\ 0 & 0 
    \end{bmatrix}\!\in\!T_\g\SE,
\end{aligned}\end{equation}
where $\hat{\eta}_2 \in \so$ and $T_\g \SE$ represent tangent space of $\SE$ at configuration $\g$.
Expressing the error function \eqref{eq:err_fun_SE3} as
\begin{equation*}\begin{aligned}
    \Psi(g,g_d) = \dfrac{1}{2}\tr\begin{bmatrix}
    2I \!-\! R^T R_d \!-\! R_d^T R & \ast\\
    \ast & (\pvec \!-\! \pvecd)^T(\pvec \!-\! \pvecd)
    \end{bmatrix}\!,
\end{aligned}\end{equation*}
where we utilize ~$\ast$ to denote the content of irrelevant matrix elements.
Its perturbation can be obtained by
\begin{align} \label{eq:perturbation_of_error_metric}
    \delta \Psi(\g,\gd)&=\!\dfrac{1}{2}\tr\!\begin{bmatrix}\!
    -\delta R^T R_d\!-\!R_d^T \delta R & \ast \nonumber
    \\
    \ast &  2(\pvec\!-\!\pvecd)^T \delta \pvec
    \end{bmatrix} \nonumber \\
    &=\!\dfrac{1}{2}\tr\!\begin{bmatrix}\!
    -\hat{\eta}_2^T R^TR_d\!-\!R_d^T R \hat{\eta}_2 & \ast
    \\
    \ast &  2(\pvec\!-\!\pvecd)^T R \eta_1 
    \end{bmatrix}\nonumber \\
    &= -\tr(R_d^T R \hat{\eta}_2) + (\pvec\!-\!\pvecd)^T R \eta_1 \nonumber \\
    &= (R_d^T R - R^T R_d)^\vee \cdot \eta_2 + (\pvec - \pvecd)^T R \eta_1.
\end{align}
%
We now define the \emph{position} error vector $\eg$ 
as
\begin{equation}\begin{aligned}\label{eq:eg}
    \eg = \begin{bmatrix}
    e_p \\ e_R
    \end{bmatrix} = \begin{bmatrix}
    R^T(\pvec - \pvecd) \\
    (R_d^T R - R^T R_d)^\vee 
    \end{bmatrix} \in \mathbb{R}^6,
\end{aligned}\end{equation}
so that $\delta \Psi(\g,\gd)\!=\!\eg ^T \eta$ and
\begin{equation}\begin{aligned}
    \hat{e}_g = \begin{bmatrix}
    R_d^T R - R^T R_d & R^T (\pvec - \pvecd) \\
    0 & 0 
    \end{bmatrix} \in \se.
\end{aligned}\end{equation}
As suggested in \cite{lee2010geometric}, the tangent vectors $\dot{\g}\!\in\!T_\g\SE$ and $\dot{\g}_d\!\in\!T_{\gd}\SE$ cannot be directly compared since they are not in the same tangent spaces. 
The desired velocity vector $V_d^b$ defined in $\gd$ can be translated into the vector $V_d^*$ in $\g$ using the following formula (See Ch 2.4 of \cite{murray1994mathematical} for the details of these steps): 
\begin{equation}\begin{aligned}
    \hat{V}_d^* = \g_{ed} \hat{V}_d^b \g_{ed}^{-1}, \text{ where } \g_{ed}\!=\!\g^{-1}\gd.
\end{aligned}\end{equation}
Additionally, it can be further represented as
\begin{equation}\begin{aligned}\label{eq:Ad_velo}
    V_d^* = \text{Ad}_{\g_{ed}}V_d^b, \text{ with } \text{Ad}_{\g_{ed}} = \begin{bmatrix}
   R_{ed} & \hat{\pvec}_{ed} R_{ed} \\ 0 & R_{ed}
    \end{bmatrix},
\end{aligned}\end{equation}
where $\text{Ad}_{\g_{ed}}: \mathbb{R}^6 \to \mathbb{R}^6$ is an Adjoint map, $R_{ed}\! =\! R^T R_d$, and $p_{ed} \!=\!-R^T(p\! -\! p_d)$. 
%
Based on this, we define the \emph{velocity} error vector $\ev$ as 
\begin{equation}\begin{aligned} \label{eq:ev}
    \ev\!=\!\underbrace{\begin{bmatrix}
    v^b \\ w^b \end{bmatrix}}_{V^b} \!-\! \underbrace{\begin{bmatrix}
     R^TR_d v_d\!+\!R^T R_d \hat{\omega}_d R_d^T (\pvec\!-\!\pvecd)\\
     R^T R_d \omega_d \end{bmatrix}}_{V_d^*} \!=\! \begin{bmatrix}
    e_v \\ e_\Omega \end{bmatrix}\!.
\end{aligned}\end{equation}
It is also worth noting that
\begin{align}
    \dot{\g} \!-\! \dot{\g}_d(\gd^{-1}\g)&\!=\! \g \hat{V}^b \!-\! \gd \hat{V}_d \gd^{-1} \g \!=\! g(\hat{V}^b \!-\! \g^{-1}\gd \hat{V}_d \gd^{-1}\g) \nonumber\\
    &= \g \hat{e}_V \in T_\g \SE,
\end{align}
which means that the velocity error vector $e_V$ can be utilized to represent the error on the tangent space at $\g$.

\subsection{Impedance Control as a Dissipative Control} \vspace{-5pt}
We consider the impedance control problem as a dissipative control design. 
Suppose that we have a Lyapunov function in a form that is similar to the total mechanical energy of the system
\begin{equation}\begin{aligned} \label{eq:lyapunov_function}
    V(t, q, \dot{q}) = K(t, q, \dot{q}) + P(t,q), 
\end{aligned}\end{equation}
where $K$ and $P$ are the kinetic and potential energy components of the Lyapunov function. Note that using total mechanical energy as a Lyapunov function is quite standard, as first proposed by \cite{koditschek1989application}.
A desired property of the Lyapunov function is a dissipativity, i.e., $\dot{V}(t, q, \dot{q})\!\leq\!0$. 
The kinetic energy component of the Lyapunov function can be  defined as follows
\begin{equation}\begin{aligned}
    K(t, q,\dot{q}) = \tfrac{1}{2}\ev^T \tilde{M} \ev,
\end{aligned}\end{equation}
where $\tilde{M}(q)$ is defined in \eqref{eq:robot_dynamics_eef} and $\ev$ is defined in \eqref{eq:ev}.

One easy option for the potential component of the Lyapunov function is to  the error function $\Psi(\g,\gd)$ in \eqref{eq:err_fun_SE3} multiplied by a scalar. 
However, recent variable impedance control techniques require multiple and varying compliance gains along different coordinates \cite{zhang2021learning, martin2019variable}. 
Therefore, we define a weighted version of the error function \eqref{eq:err_fun_SE3} as our potential function 
\begin{equation}\begin{aligned}\label{eq:potential_function}
    P(t,q) &= \tfrac{1}{2}\tr\left(\psi_k(\g,\g_d)^T\psi_k(\g,\g_d)\right)\!,\text{ where}\\
    \psi_k(\g,\g_d) &= \begin{bmatrix}
    \sqrt{K_R}(I -R_d^TR) & - \sqrt{K_p} R_d^T (\pvec - \pvecd)\\
    0 & 0 
    \end{bmatrix}\!,
\end{aligned}\end{equation}
$K_R\! \in \!\mathbb{R}^{3\times3}$, and $K_p\!\in\!\mathbb{R}^{3\times3}$ are symmetric positive definite stiffness matrices for orientation and translation, respectively.
Then, using the fact that $\tr(A)\!=\!\tr(A^T)$, leads to
\begin{equation}\label{eq:potential_function2} \begin{aligned} 
    P(t,q)\! =\! \tr\!\left(K_R(I\!-\!R_d^TR)\right)\! +\! \tfrac{1}{2}(\pvec\!-\! \pvecd)^T R_d K_p R_d^T (\pvec\!-\!\pvecd).
\end{aligned}\end{equation}
The potential function \eqref{eq:potential_function2} is the second version of the multiple $\SE$ error function alternatives presented in Table.~1 of \cite{bullo1999tracking}, and it can be shown that it is left-invariant.
Using the error vector $e_g$ and $e_V$, an intuitive impedance control law that resembles conventional impedance controllers can be formulated as follows:
\begin{equation}\begin{aligned} \label{eq:intuitive_impedance}
    \tilde{T} = \tilde{M} \dot{V}_d^* + \tilde{C} V_d^* + \tilde{G} - K_g \eg - K_d \ev
\end{aligned}\end{equation}
where $K_g = \text{blkdiag}(K_p, K_R)$. Here, the $K_g \eg$ term can be interpreted as the spring force in $T_g^*\SE$, i.e., $f_g = K_g e_g$.
However, the control action from \eqref{eq:intuitive_impedance} is not in the gradient direction of \eqref{eq:lyapunov_function}. 
Therefore, as shown in \cite{bullo1999tracking}, the elastic force $f_g$ induced from the potential function \eqref{eq:potential_function}, such that $f_g\!\in\! T_g^*\SE$, should be utilized and is given as follows:
\begin{equation}\begin{aligned} \label{eq:geometric_impedance}
    f_g = \begin{bmatrix}
    f_p \\ f_R
    \end{bmatrix} =\begin{bmatrix}
    R^T R_d K_p R_d^T (\pvec - \pvecd)\\
    (K_R R_d^T R - R^T R_d K_R)^\vee
    \end{bmatrix}
\end{aligned}\end{equation} 
The derivation of \eqref{eq:geometric_impedance} is presented in the Appendix~\ref{Sec:A}.
Using \eqref{eq:geometric_impedance}, a geometric impedance control law on $\SE$ is given by
\begin{equation}\begin{aligned}\label{eq:control_law}
    \tilde{T} = \tilde{M} \dot{V}_d^* + \tilde{C} V_d^* + \tilde{G} - f_g - K_d \ev.
\end{aligned}\end{equation}
The proposed control design is based on the following assumption.
\begin{assum} \label{assum:2}
    The end-effector of the manipulator and the desired trajectory lies in the reachable set $\mathcal{R}$, i.e., $p(\theta)\!\in\!\mathcal{R}\!=\!\{ p(\theta) \;|\; \forall\theta\!\in\!\mathcal{S}\}\!\subset\!\mathbb{R}^3$. 
    The desired trajectory is also continuously differentiable.
\end{assum}
The dissipative property of the impedance controller can be obtained by
\begin{equation}\begin{aligned} \label{eq:dissipative_property}
    \dot{V} = -\ev^T K_d \ev,
\end{aligned}\end{equation}
with the symmetric positive definite damping matrix  $K_d\!\in\!\mathbb{R}^{6\times6}$.
%
In order to obtain the control law, the time derivative of the Lyapunov function is considered as follows
\begin{equation}\begin{aligned} \label{eq:time_derivative_Lyapunov}
    \dot{V} = \dot{K} + \dot{P}.
\end{aligned}\end{equation}
The first term of the right hand side of \eqref{eq:time_derivative_Lyapunov} is calculated by using \eqref{eq:Ad_velo}-\eqref{eq:ev} as
\begin{equation}\begin{aligned} \label{eq:time_derivative_kinetic}
    \dot{K} &= \ev^T \tilde{M}(q) \dot{e}_V + \frac{1}{2}\ev^T \dot{\tilde{M}}(q) \ev\\
    \text{with }\dot{e}_V &= \dot{V}^b \!-\! \left(\tfrac{d}{dt}\text{Ad}_{\g_{ed}}\right)V_d^b \!-\! \text{Ad}_{\g_{ed}}\dot{V}^b_d = \dot{V}^b \!-\! \dot{V}_d^*,
\end{aligned}\end{equation}
\begin{equation*}\begin{aligned}
    \frac{d}{dt}\text{Ad}_{\g_{ed}} &= \begin{bmatrix}
    \dot{R}_{ed} & \hat{\dot{p}}_{ed}R_{ed} + \hat{p}_{ed}\dot{R}_{ed} \\
    0 & \dot{R}_{ed}
    \end{bmatrix}\!,\\
    \dot{R}_{ed} &= -\hat{\omega}R^TR_d + R^T R_d \hat{\omega}_d,\\
    \dot{p}_{ed}& = \hat{\omega}R^T(\pvec - \pvecd) - v^b + R^TR_d v_d.
\end{aligned}\end{equation*}
We define the time derivative of the desired velocity
$\left. \dot{V}_d^*\!\triangleq\!\left(\tfrac{d}{dt}\text{Ad}_{\g_{ed}}\right)V_d^b\!+\!\text{Ad}_{\g_{ed}}\dot{V}^b_d \right.$, so that $\dot{e}_V\!=\!\dot{V}^b\!-\!\dot{V}_d^*$.
Then, by letting $\tilde{T}_e\!=\!0$ for control law derivation, equation~\eqref{eq:time_derivative_kinetic} can be summarized as 
\begin{equation}\begin{aligned} \label{eq:delta_K}
    \dot{K} = \ev^T &\left(\tilde{T} - \tilde{C}V^b - \tilde{G} - \tilde{M}\dot{V}_d^* - \tfrac{1}{2}\dot{\tilde{M}}\ev \right).
\end{aligned}\end{equation}
For the potential energy counterpart, it can be shown that 
\begin{equation}\begin{aligned} \label{eq:time_derivative_potential}
    \dot{P} = f_g ^T e_V.
\end{aligned}\end{equation}
The derivation of \eqref{eq:time_derivative_potential} is shown 
in Appendix~\ref{Sec:A}.

The following Lemma turns out to be useful when deriving the control law.
\begin{lem} \label{lem:1}
    $\dot{\tilde{M}}\!-\!2\tilde{C}$ is skew-symmetric.
\end{lem}
\begin{pf}
    See Chap. 3.5 of \cite{lewis2003robot}.
\end{pf}
The following theorem shows that the proposed control law \eqref{eq:control_law} satisfies the dissipative property of the impedance controller \eqref{eq:dissipative_property}.
\begin{thm}
    Suppose assumptions~\ref{assum:1} and \ref{assum:2} hold true.
    Consider a robotic manipulator with the dynamic equation of motions given by \eqref{eq:robot_dynamics} and energy-based Lyapunov function candidate \eqref{eq:lyapunov_function}.
    Then, the Lyapunov function of the closed-loop system with the control law \eqref{eq:control_law} satisfies dissipative property \eqref{eq:dissipative_property}, when $T_e = 0$.
\end{thm} \vspace{-8pt}
\begin{pf} 
    Plugging in the control law \eqref{eq:control_law} into \eqref{eq:delta_K} and plugging in the resulting term into the Lyapunov \eqref{eq:time_derivative_Lyapunov}, and using the velocity error \eqref{eq:ev} as well as {\it Lemma \ref{lem:1}} leads to
    \begin{align} \label{eq:deri_Lyapu}
        \dot{V}&= \ev^T \left(\tilde{M} \dot{V}_d^* \!+\! \tilde{C}V_d^* \!-\! K_d \ev -\tilde{C}V^b \!-\! \tilde{M} \dot{V}_d^*
        \!-\! \tfrac{1}{2}\dot{\tilde{M}}\ev \right) \nonumber \\
        &= \ev^T \left(\tfrac{1}{2}\dot{\tilde{M}} - \tilde{C} - K_d \right) \ev = - \ev^T K_d \ev \leq 0,
    \end{align}
which shows that the closed-loop system is at least stable in the sense of Lyapunov (SISL), and shows proper impedance behavior as proposed in \cite{kronander2016stability}.
\end{pf} \vspace{-5pt}
The asymptotic stability of closed loop system using a strict Lyapunov analysis as presented in the following Theorem~\ref{thm:2}. 
%
%
\begin{thm} \label{thm:2}
    Suppose the assumptions~\ref{assum:1} and ~\ref{assum:2} hold true.
    Consider a robotic manipulator with dynamic equations of motion given by \eqref{eq:robot_dynamics}, energy-based Lyapunov function \eqref{eq:lyapunov_function}, and the control law \eqref{eq:control_law}.
    Then, when $T_e\! =\! 0$, the equilibrium point $\left.g(t)\! =\! g_d(t)\right.$, such that $f_g \!=\! 0$, of the closed-loop system is asymptotically stable in the reachable set $\mathcal{R}$.
\end{thm}
\begin{pf}
    First, the time derivative of $f_g = [f_p^T, f_R^T]^T$ term is calculated as follows.
    \begin{align}\hspace{-5pt} \label{eq:f_g_dot}
        \dot{f}_g \!&=\! \begin{bmatrix}
        R^TR_d K_p R_d^T R & \hat{f}_p \\
        0 & \tr(R^T R_d K_R) I\! -\! R^T R_d K_R \end{bmatrix} 
        \begin{bmatrix}
        e_v \\ e_\Omega \end{bmatrix} \nonumber\\ 
        &\triangleq\! B_K(g,g_d) e_V\!.
    \end{align}
    We denote $B_K(g,g_d) = B_K$ for the compactness of the notation. The derivation of \eqref{eq:f_g_dot} is presented in the Appendix~\ref{Sec:A}.
    By applying the control law \eqref{eq:control_law} into \eqref{eq:robot_dynamics_eef} with $\tilde{T}_e\!=\!0$, the error dynamics is  
    \begin{equation}\begin{aligned} \label{eq:error_dynamics}
        \dot{f}_g &= B_K\ev \\
        \tilde{M} \dot{e}_V &= -\tilde{C}e_V - f_g - K_d \ev.
    \end{aligned}\end{equation}
    %
    %
    Inspired by \cite{bullo1995proportional} and as suggested in \cite{lee2010geometric}, a Lyapunov function candidate $V_1$for  stability analysis can defined as follows:
    \begin{equation}\begin{aligned}
        V_1 = V + \varepsilon f_g^T \ev, \quad \varepsilon > 0.
    \end{aligned}\end{equation}
    The time-derivative of $V_1$ using the Lyapunov function \eqref{eq:deri_Lyapu} the error dynamics \eqref{eq:error_dynamics} is 
    \begin{equation}\begin{aligned}
        \dot{V}_1 &= - \ev^T K_d \ev + \varepsilon \dot{f}_g^T \ev + \varepsilon f_g^T \dot{e}_V\\
        &= -\ev^T (K_d \!-\! \varepsilon B_K) \ev - \varepsilon f_g^T \tilde{M}^{-1}f_g\\
        &\hspace{12pt}- \varepsilon f_g^T\tilde{M}^{-1}\left(\tilde{C} + K_d\right) \ev 
    \end{aligned}\end{equation}
    The matrix $B_K$ is bounded by some constant $a$, i.e., $\left.\|B_K\|\!\leq\!a\right.$
    This is because the reachable set is bounded, resulting in boundedness of $f_p$, and the term $(\tr(R^T R_d K_R) I\! -\! R^T R_d K_R)$ is bounded as similarly as proposed in \cite{lee2010geometric}.
    Therefore,
    \begin{align}
        \dot{V}_1 &\leq - \evn^2(\lambda_{\min}(K_d) - \varepsilon a) -\varepsilon \tfrac{1}{\lambda_{\max}(\tilde{M})}\|f_g\|_2^2 \nonumber\\
        &\hspace{12pt}- \varepsilon \left(\tfrac{\|\tilde{C}\| + \lambda_{\min}(K_d)}{\lambda_{\max}(\tilde{M})} \right)\|f_g\|_2\;\evn  \\
        &= -\begin{bmatrix}\|f_g\|_2 & \evn\end{bmatrix} \begin{bmatrix}
        \varepsilon c & \varepsilon b \\
        \varepsilon b & k_d\!-\!\varepsilon a
        \end{bmatrix} \begin{bmatrix}
        \|f_g\|_2 \\ \evn
        \end{bmatrix} \triangleq -z^T Q_\varepsilon z \nonumber
    \end{align}
    where $z\!=\!\left[\|f_g\|_2, \evn\right]^T\!\in\!\mathbb{R}^2$, $\lambda_{\min}(\cdot)$ and $\lambda_{\max}(\cdot)$ denotes the minimum and maximum eigenvalue of the matrix, respectively, $k_d = \lambda_{\min}(K_d)$, and
    \begin{equation*}\begin{aligned}
        b \!=\! \tfrac{\|\tilde{C}\| + \lambda_{\min}(K_d)}{2\lambda_{\max}(\tilde{M})},\; c \!=\! \tfrac{1}{\lambda_{\max}(\tilde{M})},\; Q_\varepsilon \!=\! \begin{bmatrix}
        \varepsilon c & \varepsilon b \\
        \varepsilon b & k_d \!-\!\varepsilon a
        \end{bmatrix}\!\in\!\mathbb{R}^{2\times 2}\,.
    \end{aligned}\end{equation*}
    With sufficiently large gains $K_d, K_p, K_R$ and/or taking sufficiently small $\varepsilon$, one can make $Q_\varepsilon$ positive definite. In particular, taking
    \begin{equation}\begin{aligned}
        \varepsilon < \min \left\{ \tfrac{k_d}{|a-c|}, \tfrac{k_d c}{b^2 + ac} \right\}
    \end{aligned}\end{equation}
    renders $Q_\varepsilon$ positive definite, which completes the proof.
\end{pf}
\subsection{Impedance Control as an Exact Compensation Tracking Control Law} \vspace{-5pt}
We now propose another  geometric impedance control law utilizing the controller design techniques suggested by \cite{sadegh1990stability, slotine1987adaptive} in order to attain a more robust stability result. First, a reference velocity $\bar{V}_d$ and a reference acceleration $\dot{\bar{V}}_d$ is defined by
\begin{equation} \label{eq:ref_vel}
    \begin{aligned}
        \bar{V}_d &= V_d^* - \lambda_g f_g \\
        \dot{\bar{V}}_d &= \dot{V}_d^* - \lambda_g \dot{f}_g = \dot{V}_d - \lambda_g B_k e_V,
    \end{aligned}
\end{equation}
where $\lambda_g \in \mathbb{R}_+$ is a positive scalar gain, but it is not restricted to the scalar. 
Then, a reference velocity error term $\bar{e}_V$ can be defined accordingly as follows:
\begin{equation} \label{eq:ref_vel_err}
    \begin{aligned}
        \bar{e}_V = V^b - \bar{V}_d = V^b - V_d^* + \lambda_g f_g = e_V + \lambda_g f_g
    \end{aligned}
\end{equation}
Using \eqref{eq:ref_vel_err}, we propose a second version of the geometric impedance control law in the following formulation:
\begin{equation}\label{eq:control_law_2}
    \begin{aligned}
        \tilde{T} = \tilde{M} \dot{\bar{V}}_d + \tilde{C} \bar{V}_d - f_g - K_d \bar{e}_V + \tilde{G}.
    \end{aligned}
\end{equation}
The stability proof for the geometric impedance control law 2 is presented in the following theorem.
\begin{thm}
    Suppose the assumptions~\ref{assum:1} and ~\ref{assum:2} hold true.
    Consider a robotic manipulator with dynamic equations of motion given by \eqref{eq:robot_dynamics} and the control law \eqref{eq:control_law_2}.
    Then, when $T_e\! =\! 0$, the equilibrium point $\left.g(t)\! =\! g_d(t)\right.$, such that $f_g \!=\! 0$, of the closed-loop system is asymptotically stable in the reachable set $\mathcal{R}$.
\end{thm}
\begin{pf}
    We start the proof by defining the error dynamics under the control law \eqref{eq:control_law_2}.
    \begin{equation} \label{eq:err_dynamics_2}
        \begin{aligned}
            \dot{f}_g &= B_K (\bar e_V - \lambda_g  f_g)\\
            \tilde{M} \dot{\bar{e}}_V &= - \tilde{C} \bar{e}_V - f_g - K_d \bar{e}_V 
        \end{aligned}
    \end{equation}
    The energy-function-like Lyapunov function candidate $W(t,q,\dot{q})$ is defined by 
    \begin{equation} \label{eq:lyapunov_function_2}
        \begin{aligned}
            W(t,q,\dot{q}) &= \bar{K} (t,q,\dot{q}) + P(t,q) \\
            \bar{K}(t,q,\dot{q}) &= \tfrac{1}{2} \bar{e}_V^T \tilde{M} \bar{e}_V,
        \end{aligned}
    \end{equation}
    The time-derivative of the Lyapunov function $\dot{W}$ is evaluated based on \eqref{eq:err_dynamics_2} and Lemma~\ref{lem:1} as follows:
    \begin{equation}
        \begin{aligned}
            \dot{W} &= \dot{\bar{K}} + \dot{P}, \quad \text{where}\\
            \dot{\bar{K}} &= \bar{e}_V^T\left(\tfrac{1}{2}\dot{\tilde{M}}\bar{e}_V - \tilde{C} \bar{e}_V - f_g - K_d \bar{e}_V \right)\\
            &= -\bar{e}_V^T f_g - \bar{e}_V^T K_d \bar{e}_V\\
            \dot{P} &= f_g^T {e}_V = f_g^T(\bar{e}_V - \lambda_g f_g)
            = \bar{e}_V^T f_g - \lambda_g f_g^T f_g.
        \end{aligned}
    \end{equation}
    Therefore, by using $f_g$ and $\bar{e}_V$ as the states of the Lyapunov function $W$,
    \begin{equation}
        \begin{aligned}
            \dot{W} = - \bar{e}_V^T K_d \bar{e}_V - \lambda_g f_g^T f_g < 0
        \end{aligned}
    \end{equation}
    which shows negative definiteness of $\dot{W}$, implying asymptotic stability.
\end{pf}
We present some comments regarding the second version of the impedance control law in the following remark. 
\begin{rem} Comments on control law \eqref{eq:control_law_2}
\begin{enumerate}
    \item When control law \eqref{eq:control_law_2} is utilized, the effective stiffness of the end-effector is changed. Therefore, one should choose the gains $K_p$ and $K_R$ carefully considering the effective stiffness.
    \item When $\lambda_g \to 0$, the control law \eqref{eq:control_law_2} becomes identical to \eqref{eq:control_law}.
    \item The stability of the \eqref{eq:control_law} is restricted by the selection of trajectories and gains because of its dependence on matrix norms such as $\|B_K\|$ and $\|\tilde{C}\|$, i.e., $\varepsilon$ value sometimes needs to be small. On the other hand, for \eqref{eq:control_law_2}, $\lambda_g$ is not dependent on these values, resulting in more generalized stability results.
\end{enumerate}
\end{rem}

\section{Simulation Result} \label{Sec:4}
We utilize the UR5e robot model implemented in Matlab~R2021a using \cite{corke2002robotics}.
%
%
As a benchmark, we use a conventional impedance control from \cite{robosuite2020, ochoa2021impedance, shaw2022rmps}, which does not consider geometrical aspects, while allowing  comparisons under similar initial conditions and gains to the controller presented in this paper. 
Specifically, the control law utilized for the benchmark is as follows:
\begin{equation} \label{eq:conventional_impedance}
    \begin{aligned}
        \tilde{T}^s = \tilde{M}^s\dot{V}_d + \tilde{C}^s V + \tilde{G}^s - K_g e_g^s - K_d {e}_V^s
    \end{aligned}
\end{equation}
where $\tilde{M}^s$, $\tilde{C}^s$, and $\tilde{G}^s$ can be obtained by replacing $J_b$ by $J_s$, a spatial Jacobian matrix, from \eqref{eq:robot_dynamics_eef}, and $\tilde{T}^s = J_s^{-T}T$. In addition, the spatial error vector $e_g^s$ is defined as \cite{robosuite2020, ochoa2021impedance, shaw2022rmps}
\begin{equation}
    \begin{aligned}
         e_g^s &= [
         (e_p^s)^T, \; (e_R^s)^T
         ]^T, \quad \text{where}\\
        e_p^s &= x \!-\! x_d, \quad e_R^s = (r_{d_1}\! \times\! r_1 + r_{d_2}\! \times \! r_2 + r_{d_3}\! \times\! r_3),
    \end{aligned}
\end{equation}
with $R = [r_1, r_2, r_3]$ and $R_d = [r_{d_1}, r_{d_2}, r_{d_3}]$. Likewise, the spatial velocity error $e_V^s$ is
\begin{equation}
    \begin{aligned}
        e_V^s = \begin{bmatrix}
            v^s \\ \omega^s
        \end{bmatrix} - \begin{bmatrix}
            \dot{p}_d \\ \omega_d^s
        \end{bmatrix} = V^s - V_d^s,
    \end{aligned}
\end{equation}
where the spatial velocity is obtained by $V^s = J_s \dot{q}$. Note that this controller is very similar to the one proposed in \cite{caccavale1999six}, where the only difference is the representation of the rotational error $e_R^s$. 

\begin{figure}[t!]
    \vspace{-5pt}
    \hspace{-8pt}
    \input{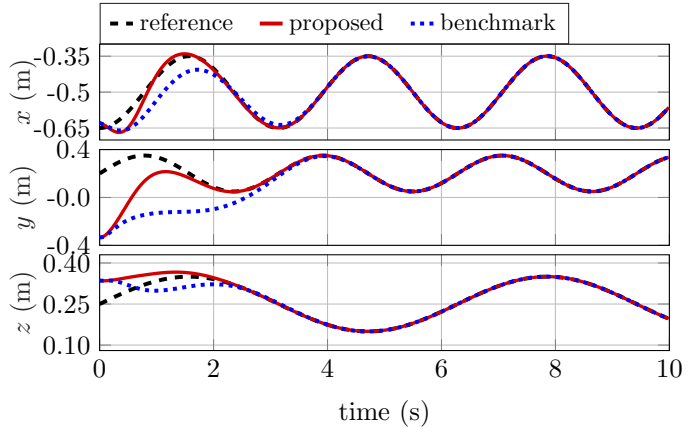}
    \vspace{-20pt}
    \caption{The trajectory tracking results in $x$, $y$, and $z$ coordinates for the proposed and benchmark approach are plotted.}
    \label{fig:comparison_xyz_tracking}
\end{figure}

\begin{figure}
    \vspace{-5pt}
    \hspace{-8pt}
    \input{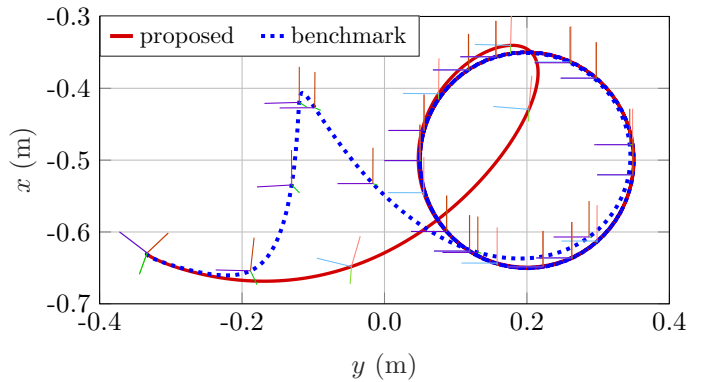}
    \vspace{-20pt}
    \caption{Trajectory results in a bird-eye-view for $t = 0 \sim 10 (\mathrm{s})$ with end-effector axes attached are plotted.}
    \label{fig:2d_traj}
\end{figure}

The simulations are conducted for a dynamic trajectory with a rapid motion tracking scenario. We present a simulation result with the first version of the geometric control law \eqref{eq:control_law} since it enables a fair comparison with the benchmark controller. Additionally, we set the same desired impedance $K_p$, $K_R$, and $K_d$ to both controllers and also let both controllers start at the same initial condition. The desired impedance gains are set as
\begin{equation*}\begin{aligned}
    K_p \!=\! \text{diag}\left([200,60,80]\right),K_R \!= \! \text{diag}\left([10,30,100]\right), K_d \!=\! k_d I
\end{aligned}\end{equation*}
where $k_d\!=\!50$. The initial point in the simulation is selected as
\begin{equation*}\begin{aligned}
    q(0) &= [0.2,-0.5,0.4,0.6,-0.5,0.2]^T,\\
    \dot{q}(0) &= [0,0,0,0,0,0]^T.
\end{aligned}\end{equation*}
The desired trajectories are
\begin{equation} \begin{aligned}
    p_d(t) = \begin{bmatrix}
    -0.5 - 0.15 \cos{2t}\\
    0.2 + 0.15 \sin{2t}\\
    0.25 + 0.1 \sin{t}
    \end{bmatrix}, \quad R_d(t) = \begin{bmatrix}
        1 & 0 & 0\\
        0 & 0 & -1 \\
        0 & 1 & 0
    \end{bmatrix}
\end{aligned} \end{equation}
The Cartesian trajectory tracking results are plotted in Fig.~\ref{fig:comparison_xyz_tracking}.
The trajectory results in $x\text{-}y$ plane are shown in Fig.~\ref{fig:2d_traj}
The RMS values for the performance metrics are shown in Tab.~\ref{table:rms_of_errors_tracking}. 
Both the proposed control and benchmark control showed perfect trajectory tracking performance once after they were converged. 
However, the proposed control law improved performance in reducing the initial errors, especially in the translational part, as seen in Tab.~\ref{table:rms_of_errors_tracking} and Fig.~\ref{fig:2d_traj}. This is mainly because the benchmark control tries to reduce the similar potential function separately in translational and orientational direction, while the proposed control approach reduces the translational and orientational error simultaneously.
\begin{table}[t!]
  \caption{Performance metrics in RMS values for the dynamic trajectory tracking scenario}
  \label{table:rms_of_errors_tracking}
  \scriptsize
  \centering%
  \begin{tabu}[width=\columnwidth]{%
      c| c | c
    }
    \toprule
    Performance Metrics & Benchmark & Proposed\\
    \midrule
    $\text{RMS}(x(t) - x_d(t))$ & 0.0317 & 0.0137\\
    $\text{RMS}(y(t) - y_d(t))$ & 0.1992 & 0.1256\\
    $\text{RMS}(z(t) - z_d(t))$ & 0.0183 & 0.0178\\
    $\text{RMS}(P(t,q))$, \eqref{eq:potential_function} & 6.5693 & 6.3624\\
    $\text{RMS}(V(t,q,\dot{q}))$, \eqref{eq:lyapunov_function} & 7.2622 & 6.6556\\
    \bottomrule
  \end{tabu}
\end{table}

The Matlab code for the implementation of the controller proposed in this paper is provided in \url{https://github.com/Joohwan-Seo/Geometric-Impedance-Control-Public}.\\
Few remarks on the actual implementation are provided.
\begin{rem} 
    From the implementation perspective, the normal PD-like control with gravity compensation given by
    $\tilde{T}\!=\!-f_g(e_g) \!-\! K_d \ev + \tilde{G}$
    can be utilized when the Jacobin matrix is numerically near singular. Similarly, $\tilde{T}\!=\!-K_g e_g \!-\! K_d \ev + \tilde{G}$ for the intuitive impedance control.
    This implies that even without dynamic parameters and solving inverse kinematics, we can solve the end-effector positioning problem, which has already been emphasized in an earlier work 
    \cite{hogan1985impedance}.
    %
\end{rem}

\section{Conclusion and Future Works} \label{Sec:5}
A novel impedance control law considering the geometric structure of the manipulator is proposed in this paper.
The left-invariant error metric is first defined, and the geometrically consistent error vectors are derived.
Based on the error vectors, the geometric impedance control rule is obtained from the dissipative property of the energy function. 
The stability analysis shows that the closed-loop system is asymptotically stable.
In the simulation, 
we showed that the proposed control outperformed the conventional control in trajectory tracking.

In future work, we will combine the proposed geometric impedance control with reinforcement learning to solve robotic manipulation task problems.
Since the controller is now defined appropriately, we expect an improvement in performance compared to state-of-the-art methods. We will also show that task encoding based on a geometrically consistent error vector is useful.
Above that, we will also research the solution of robotic tasks where the tasks can be better encoded as velocity fields instead of desired trajectories, i.e., a closed-loop system lying on the desired manifold.
%


\bibliography{biblio} 

\appendix
\section{}
\label{Sec:A}

\subsection{A derivation of geometric elastic force $f_g$ \eqref{eq:geometric_impedance}}
\vspace{-8pt}
Similar to \eqref{eq:perturbation_of_error_metric}, the potential function \eqref{eq:potential_function} is perturbed as follows:
\begin{align*}
    \delta P = -\tr(K_R R_d^T R \hat{\eta}_2) + (p - p_d)^T R_d K_p R_d^T R \eta_1
\end{align*}
The first term in the equation above is further evaluated as
\begin{align*}
    -\tr(K_R R_d^T R \hat{\eta}_2) &=-\tfrac{1}{2}\tr\left((K_R R_d^T R - R^T R_d K_R) \hat{\eta}_2\right) \\
    &= (K_R R_d^T R - R^T R_d K_R)^\vee \cdot \eta_2 = f_R^T \eta_2,
\end{align*}
where $-\tfrac{1}{2}\tr(\hat{x}\hat{y}) = x^T y$ is utilized.
The second term can be similarly represented using $f_p = R^T R_d K_p R_d^T (p - p_d)$.
The perturbed potential function can be notated as follows:
\begin{align*}
    \delta P = f_p^T \eta_1 + f_R^T \eta_2 = \begin{bmatrix}
    f_p^T & f_R^T
    \end{bmatrix} \begin{bmatrix}
    \eta_1 \\ \eta_2
    \end{bmatrix} = f_g^T \eta.
\end{align*}
Since $\eta$ is arbitrary, we can represent an elastic force in $\SE$ with respect to the potential function \eqref{eq:potential_function} as $f_g$.

\vspace{-6pt}
\subsection{A derivation of the time-derivative of the potential energy part \eqref{eq:time_derivative_potential}}
\vspace{-8pt}
The time-derivative of the potential function is
\begin{align*}
    \dfrac{dP}{dt} = \dfrac{1}{2}\tr(\dot{\psi}_k^T\psi_k) + \dfrac{1}{2}\tr(\psi_k^T \dot{\psi}_k) = \tr(\psi_k^T \dot{\psi}_k)
\end{align*}
Where,
\begin{align*}
    &\dot \psi_k = 
    \dfrac{d\psi_k}{dt} 
    = \begin{bmatrix}
    -\sqrt{K_R}R_d^T R \hat{e}_\Omega & m_{12}\\
    0 & 0
    \end{bmatrix},\mbox{with }\\
     &m_{12} = \sqrt{K_p} \hat{\omega}_d R_d^T (\pvec - \pvecd) - \sqrt{K_p} R_d^T Rv + \sqrt{K_p} v_d.
\end{align*}
Which leads to the form of
\begin{align*}
    \dfrac{dP}{dt} = \tr\left(\begin{bmatrix}
    a_{11} & \ast \\
    \ast & a_{22}
    \end{bmatrix}
    \right)
\end{align*}
where $\ast$ terms are not of interest because of the trace operator. 
\begin{align*}
    a_{11} &= (I - R^TR_d)\sqrt{K_R}^T (-\sqrt{K_R})R_d^T R \hat{e}_\Omega  \\
    a_{22} &= -(\pvec - \pvecd)^T R_d \sqrt{K_p}^T m_{12}
\end{align*}
Using the assumption that $K_R$ is a symmetric positive definite matrix, and matrices are commutable in the trace, i.e., $\tr(AB) = \tr(BA)$,
\begin{align*}
    \tr(a_{11}) &= -\tr((I -R^T R_d )K_R R_d^T R\hat{e}_\Omega)\\
    &= - \tr(K_R R_d^TR\hat{e}_\Omega) + \tr(R^T R_d K_R R_d^T R \hat{e}_\Omega),
\end{align*}
where the second term is evaluated as
\begin{align*}
    \tr(R^T R_d K_R R_d^T R \hat{e}_\Omega) = 0
\end{align*}
from the fact that $\tr(AB) = \tr(B^T A^T) = \tr(A^T B^T) = -\tr(AB)$ with $A^T = A$ and $B^T = -B$. Then, the remaining term is
\begin{align*}
    \tr(a_{11})&= -\dfrac{1}{2}\tr(K_R R_d^T R \hat{e}_\Omega) + \dfrac{1}{2} \tr(R^T R_d K_R \hat{e}_\Omega)\\
    &= -\dfrac{1}{2}\tr((K_R R_d^T R - R^T R_d K_R) \hat{e}_\Omega) = f_R ^T \hat{e}_\Omega
\end{align*}
On the other hand, $\tr(a_{22}) = a_{22}$ can be unraveled as
\begin{align*}
    &a_{22} = -(\pvec - \pvecd)^T  R_d K_p  \hat{\omega}_d R_d^T(\pvec - \pvecd) \\
    & + (\pvec - \pvecd)^T R_d K_p R_d^T Rv - (\pvec - \pvecd)^T R_d K_p v_d\\
    &= -(\pvec - \pvecd)R_d K_p R_d^T R R^T R_d \hat{\omega}_d R_d^T(\pvec - \pvecd)  \\
    & + (\pvec - \pvecd)^T R_d K_p R_d^T R (v - R^T R_d v_d)\\
    &= (\pvec \!-\! \pvecd)^T R_d K_p R_d^T R (v \!-\! R^TR_d v_d \!-\! R^TR_d \omega_d R_d^T (\pvec \!-\! \pvecd))\\
    &= f_p^T e_v.
\end{align*}
Putting all things together,
\begin{align*}
    \dfrac{dP}{dt} = \begin{bmatrix}
    f_p^T & f_R^T
    \end{bmatrix} \begin{bmatrix}
    e_v \\ e_\Omega
    \end{bmatrix} = 
    f_g^T e_V,
\end{align*}
which is \eqref{eq:time_derivative_potential}, where $f_g$ is defined in \eqref{eq:geometric_impedance}.

\vspace{-6pt}
\subsection{Derivation of $\dot{f}_g$ \eqref{eq:f_g_dot}}
\vspace{-8pt}
The time derivative of $f_R$ is
\begin{align*}
    \dot{f}_R = \frac{d}{dt}\left(K_R R_d^T R - R^T R_d K_R\right)^\vee, \; \text{where}
\end{align*}
\begin{align*}
    &\dfrac{d}{dt}\left(K_R R_d^T R - R^T R_d K_R\right)\\
    &= K_R\left(\hat{\omega}_d^T R_d R + R_d R \hat{\omega}\right) -  \left(\hat{\omega}^TR^T R_d + R^T R_d \hat{\omega}_d\right)K_R\\
    &= K_R\left(-\hat{\omega}_d R_d R + R_d R \hat{\omega}\right) +  \left(\hat{\omega}R^T R_d - R^T R_d \hat{\omega}_d\right)K_R\\
    &= K_RR_d^TR\hat{e}_\Omega + \hat{e}_\Omega R^T R_d K_R.
\end{align*}
Therefore, 
\begin{align*}
    \dot{f}_R &= \left(K_R R_d^TR\hat{e}_\Omega + \hat{e}_\Omega R^T R_d K_R\right)^\vee \\
    &= \left(\tr(R^T R_d K_R) I - R^T R_d K_R \right) e_\Omega,
\end{align*}
where the following useful lemma is utilized.
\begin{lem}
    For $A \in \mathbb{R}^{3\times 3}$, $b \in \mathbb{R}^3$,
    \begin{align*}
        (A \hat{b} + \hat{b} A^T)^\vee = (\tr(A^T)I - A^T) b.
    \end{align*}
\end{lem}
The time derivative of $f_p$ is
\begin{align*}
    & \dot{f}_p = \frac{d}{dt}(R^T R_d K_p R_d^T (p - p_d))\\
    &= -\hat{\omega}R^T R_d K_p R_d^T (p \!-\! p_d) \!+\! R^T R_d \hat{\omega}_d K_p R_d^T(p \! -\! p_d) \\
    &- R^TR_d K_p \hat{\omega}_d R_d(p\! -\! p_d)\! +\! R^T R_d K_p R_d^T (Rv\! -\! R_d v_d) \\
    &= -\underbrace{(\hat{\omega} - R^T R_d \hat{\omega}_d R_d^T R)}_{\hat{e}_\Omega} \underbrace{R^T R_d K_p R_d^T(p \!-\! p_d)}_{f_p}\\
    &+ R^T R_d K_p R_d^T R \underbrace{(v \!-\! R^T R_d v_d \!-\! R^T R_d \hat{\omega}_d R_d(p\! -\! p_d))}_{e_v}\\
    &= -\hat{e}_{\Omega} f_p + R^T R_d K_p R_d^T R e_v = \hat{f}_p e_{\Omega} + R^T R_d K_p R_d^T R e_v.
\end{align*}
Combining with the previous result, it is shown that
\begin{align*}
    \dot{f}_g &=\! \begin{bmatrix}
    R^T R_d K_p R_d^T R & \hat{f}_p \\
    0 & \tr(R^T R_d K_R) I\! -\! R^T R_d K_R
    \end{bmatrix} \begin{bmatrix}
    e_v \\ e_\Omega
    \end{bmatrix}\\
    &= B_K(g,g_d) e_V.
\end{align*}

\end{document}